\documentclass{article}
\usepackage{spconf,amsmath,graphicx}
\usepackage{svg}
\usepackage{cite}
\usepackage[utf8]{inputenc} 
\usepackage[T1]{fontenc}    
\usepackage{hyperref}       
\usepackage{url}            
\usepackage{booktabs}       
\usepackage{amsfonts}       
\usepackage{nicefrac}       
\usepackage{microtype}      
\usepackage{bm} 
\usepackage{amsmath,amssymb,amsthm}
\usepackage{graphicx}
\graphicspath{{./figures/}}
\usepackage{subcaption}
\usepackage{enumitem}
\usepackage{tikz}
\usepackage{soul} 
\usepackage{cite}
\usepackage[ruled,vlined]{algorithm2e}
\usepackage[british]{babel}
\usepackage{hhline}
\usepackage{multirow}
\usepackage{bbm}

\newtheorem{theorem}{Theorem}
\newtheorem{lemma}{Lemma}
\renewcommand{\vec}[1]{\bm{#1}}
\DeclareMathOperator*{\argmin}{arg\,min}
\DeclareMathOperator{\sign}{sign}

\newcommand*{\trans}{^{\mkern-1.5mu\mathsf{T}}}
\newcommand*{\tran}{^{\mkern-1.5mu\mathsf{T}}}

\newcommand{\Wpm}{\vec{W}_{\pm}}
\newcommand{\Wzero}{\vec{W}_{0}}

\newcommand{\x}{{\vec{x}}}
\newcommand{\y}{{\vec{y}}}
\newcommand{\z}{{\vec{z}}}

\newcommand{\U}{{\vec{U}}}

\newcommand{\Wno}{{\vec{W}}}
\newcommand{\Q}{{\mathcal{Q}}}
\newcommand{\J}{{\mathcal{J}}}
\newcommand{\R}{\mathcal{R}}
\newcommand{\xopt}{{\vec{x}^*}}
\newcommand{\A}{{\vec{A}}}
\newcommand{\s}{{\vec{s}}}

\newcommand{\C}{{\vec{C}}}
\newcommand{\Rmat}{\vec{R}}

\newcommand{\I}{{\vec{I}}}
\newcommand{\proj}{{\vec{P}}}
\newcommand{\nullspace}[1]{{\mathcal{N}(#1)}}

\title{Bilevel learning of  $\ell_{1}$ regularizers with closed-form gradients (BLORC)}
\name{Avrajit Ghosh$^{1}$, Michael T. Mccann$^{2}$, Saiprasad Ravishankar$^{1,3}$\thanks{Emails: A. Ghosh (\textit{ghoshavr@msu.edu}), M. T. McCann (\textit{mccann@lanl.gov}), and S. Ravishankar (\textit{ravisha3@msu.edu}).}\vspace{-0.01in}}
\address{\hspace{-0.2in}$^1${Dept. of Computational Mathematics, Science and Engineering, Michigan State University, East Lansing, MI.}\\  
$^2${Theoretical Division, Los Alamos National Laboratory, Los Alamos, NM.}\\
$^3${Dept. of Biomedical Engineering, Michigan State University, East Lansing, MI.}}
\begin{document}
%
\maketitle

 \begin{abstract}

We present a method for supervised learning of sparsity-promoting regularizers, a key ingredient in many modern signal reconstruction problems.
The parameters of the regularizer are learned to minimize the mean squared error of reconstruction on a training set of ground truth signal and measurement pairs.
Training involves solving a challenging bilevel optimization problem with a nonsmooth lower-level objective.
We derive an expression for the gradient of the training loss using the implicit closed-form solution of the lower-level variational problem given by its dual problem, and
provide an accompanying gradient descent algorithm (dubbed BLORC) to minimize the loss.
Our experiments on simple natural images and for denoising 1D signals
show that the proposed method can learn meaningful operators and the analytical gradients calculated are faster than standard automatic differentiation methods.
While the approach we present is applied to denoising,
we believe that it can be adapted to a wide-variety of inverse problems with linear measurement models, thus
giving it applicability in a wide range of scenarios.         
\end{abstract}

\section{Introduction}
\label{sec:intro}


Regularized image reconstruction has been used in many applications in imaging and signal processing.
Typically, we are interested in solving the optimization problem of the form:\\
 \begin{equation}
   \hat{\x} = \underset{x}{\arg \min } \| \A \x - \y \|_2^2 + \R(\x)  
   \label{gen}
 \end{equation}
 where $\A \in \mathbb{R}^{m \times n}$ is a linear forward model or measurement matrix, $\R: \mathbb{R}^n \to \mathbb{R}$
is a regularization functional, $\x \in \mathbb{R}^n$ denotes the underlying signal that we aim to recover, and $\y \in \mathbb{R}^m$ denotes its measurements.
A prominent theme in designing the regularization functional, $\R$,
has been that of sparsity~\cite{donoho2006compressed}:
the idea that the reconstructed signal $\x$ should admit a sparse 
(having a small number of nonzero elements)
representation in some domain. While the $\ell_{0}$ ``norm" is a canonical measure of sparsity, but due to its non-convexity, its relaxed version, the $\ell_{1}$ norm is often used in practice. A common and successful approach to promoting sparsity
in signal reconstructions
is to use 
$\R(\x) = \|\Wno \x\|_1$,
where $\Wno \in \mathbb{R}^{k \times n}$
is an analysis dictionary or sparsifying transform.  While there are several choices for $\Wno$
that work well in practice
(e.g., wavelets, finite differences, 
discrete cosine transform (DCT)),
several authors have also sought to learn $\Wno$ from data,
an approach called dictionary (or transform) learning \cite{ravishankar2013learning,ravishankar2012learning,ravishankar_mr_2011}. 

Learning the regularizer in a supervised fashion has been often cast as a bilevel problem in earlier works~\cite{holler2018bilevel,kunisch_bilevel_2013,liu2019bilevel,alberti2021learning,ochs_bilevel_2015}.
One of the 
methods to solve the bilevel problem was to replace the lower-level problem by its first order optimality conditions obtained through KKT conditions and then solving the bilevel problem using an implicit differentiation method. However, sparsity-promoting regularizers typically involve nonsmooth penalties such as the $\ell_{1}$ norm.
Some works~\cite{chen_learning_2014,sprechmann_supervised_2013,yunjin2012learning,peyre_learning_2011} used relaxed versions of the $\ell_{1}$ norm to perform implicit differentiation. However, these smooth approximations introduce errors in calculating the gradients close to the non-differentiable points. 
We propose a direct approach to solving the bilevel formulation by replacing the lower-level variational formulation by an implicit  closed-form expression.  While we demonstrate supervised learning of 
transforms for denoising ($\A= \I$), our approach can be extended to general inverse problems.


Automatic differentiation packages like PyTorch~\cite{NEURIPS2019_9015} and CVXPY's Diffcp~\cite{diffcp,diffcp2019,amos2019differentiable} allow for the exact evaluation of the Jacobian of an arbitrarily complicated differentiable function, by partitioning the function into a sequence of simple operations, which are by themselves trivially differentiable. We compare our approach of deriving gradients by implicit differentiation of a  closed-form expression with the auto-differentiation approaches in terms of accuracy and time of computation. To our knowledge, no work has to date solved the bilevel sparse signal reconstruction problem by exploiting implicit closed-form expressions or without making relaxations. 
Extensive experimental results on 1D signals and  images demonstrate that our method learns meaningful transforms taking into account both the 
dataset and the task at hand, and with less runtime than the autodifferentiation methods.

\section{Bilevel formulation of Supervised learning and its analysis}

 The general bilevel sparsifying transform learning formulation for denoising  is:
\begin{equation} \label{eq:bilevel}
\begin{split}
   \underset{\Wno}{\arg \min }                     \,\Q(\Wno) = \sum_{t=1}^{T}  \frac{1}{2} \|\x_{t}^*(\Wno) - \x_t \|_2^2 
  \\
\text{s.t.} \;\; \x_{t}^*(\Wno) = \underset{x}{\arg \min} \| \x - \y_t \|_{2}^2 + ||\Wno \x||_{1}
     \end{split}
\end{equation}

The bilevel learning formulation~\eqref{eq:bilevel} consists of two problems: the upper-level  and the lower-level problem. The upper-level problem minimizes the cost  $\Q(\Wno)$ with respect to the parameters $\Wno$, where $\Q(\Wno)$ compares the ground truth $\x_{t}$ with the parameterized reconstruction output $\x_{t}^*(\Wno)$  obtained from the lower-level optimization problem. The lower-level problem is a variational reconstruction problem to obtain the parameterized reconstruction solution $\x_{t}^*(\Wno)$ from $\y_t$.

In this supervised setting, the training samples $(\x_t,\y_t)$ are given and denote a clean image and its noisy version. Our objective is to learn the transform matrix $\Wno$ so that the lower level reconstruction using this sparsity operator is as close as possible to the ground truth. It is important to note two things in the above bilevel problem: 
first, no constraint is imposed on $\Wno$ during learning; and 
second, the algorithm learns the scaling of the regularization penalty, hence a separate scalar regularization strength is not required (i.e., $\beta ||\Wno \x||_{1} = ||\beta \Wno \x||_{1}$ for $\beta \geq 0$). 
As a first step towards solving~\eqref{eq:bilevel}, we derive an implicit closed-form expression for the  lower-level problem. 

\subsection{Closed-form solution obtained by duality }
Consider the lower-level  functional 
\begin{equation}
     \label{eq:analysis}
       \J(\x, \Wno, \y) = \frac{1}{2}\| \x - \y \|_2^2 +  \| \Wno \x \|_1,
 \end{equation}
 with $\x,\y \in \mathbb{R}^n$
 and
 $\Wno \in \mathbb{R}^{k \times n}$.
It is strictly convex in $\x$
(the first term is strictly convex, while the second is convex),
and therefore, has a unique global minimizer.
Thus, 
we can write
$\x^*(\Wno) =\argmin_\x \J(\x, \Wno)$,
without the possibility of
the minimizer not existing or being a nonsingleton.
Note that although $\x^*$ depends on both $\y$ and $\Wno$, the $\y$-dependence is not relevant for this derivation and we will not include it explicitly.

A key component in deriving the closed-form expression for the minimizer of~\eqref{eq:analysis}  is that we need to know the sign pattern of  $\Wno\x^*(\Wno)$. So, the closed-form expression is an implicit equation ~\eqref{eq:closed_form}, where the reconstruction $\x^*(\Wno)$ is dependent on $\sign (\Wno \x^*)$  (where $\sign(\mathbf{\z})_{i}$ 
is defined to be -1 when $\mathbf{\z}_{i} < 0$;
0 when $\mathbf{\z}_{i} = 0$;
and 1 when $\mathbf{\z}_{i} > 0$), which itself is a function of $\x^*$. 
So, to obtain the closed-form expression, we first solve for $\x^*$ using any well-known iterative minimization algorithm like ADMM or PGD \cite{boyd_distributed_2011} and use it to compute the sign pattern. It is natural to question the approach of using an implicit closed-form equation if there exist well-known iterative minimization algorithms like ADMM and PGD to solve for  $\x^*$.
But we 
note
that our objective in this step is not to find $\x^*$, but rather to obtain a closed-form expression that allows us to take gradients with respect to $\Wno$. 
  
Let $k_{=0}$ denote the set  $\{i \in (1,2,3,..,k) \,:\, (\Wno \x^*(\Wno))_{i} =0  \}$ and $k_{\ne 0}$ denotes $\{i \in (1,2,3,..,k) \,:\,  (\Wno \x^*(\Wno))_{i}\ne 0  \}$, then we have the following theorem.
\begin{theorem}[Closed-form expression of $\argmin_\x \J(\x, \Wno)$]
Let the nonzero pattern $\s$ denote $\Rmat \sign (\Wno \x^* )$ and let $\Wzero$ and $\Wpm$ contain the rows of $\Wno$, whose indices are given by the set $k_{=0}$ and $k_{\ne0}$, respectively.
Then, the closed-form expression  for $ \x^*(\Wno) = \arg \min_{\x} \frac{1}{2}\| \x - \y \|_2^2 +  \| \Wno \x \|_1$ is obtained from Lagrangian dual analysis~\cite{tibshirani2011solution} as
\begin{equation}  
   \x^*(W)= \proj_\nullspace{\Wzero} (\y - \Wpm \tran \s ),
   \label{eq:closed_form}
\end{equation}
where $\proj_\nullspace{\Wzero} $ is the projector matrix onto the nullspace of $\Wzero$ and is given by  $\proj_\nullspace{\Wzero} = (\vec{I} - \Wzero^{\dagger} \Wzero)$, and $\Rmat$ is a row selection matrix that helps retain only the nonzero elements of the sign vector $\sign (\Wno \x^* )$.
\end{theorem}


 
\subsection{Gradient calculations}
Once, we have a closed-form expression of the lower level problem, 
hour next step is to compute the derivative of $\x^*(\Wno)$ with respect to $\Wno$
using matrix calculus~\cite{minka_old_2000}. 
 Note that in order for the gradients to exist, the sign pattern vector $\sign (\Wno \x^* )$ has to remain constant in an open set containing $\Wno$. Only then the closed-form expression for $\x^*(\Wno)$
 is valid in each region where $\sign (\Wno \x^* )$ is constant. As a brief example of this, consider the scalar denoising problem
$x^*(w) = \argmin_x \frac{1}{2}(x-y)^2 + |w x|$ and $s(w) =\sign (wx^*(w))$.
Assuming that $y\ge0$,
one can show that $x^*(w) = y-|w|$ when $y-|w| \ge 0$ and $0$ otherwise.
As a result, $s((0, y)) = 1$, $s((-y, 0)) = -1$, and $s((-\infty, -y] \cup 0 \cup [y,\infty))=0$ is piecewise constant;
a similar result holds when $y\le0$.
Thus $\Q(w)= (x^*(w) - x_{t})^{2}$ is smooth except at $w=0,-y,y$. So, there exist intervals in the domain of $w$, where $\Q(w)$ is differentiable with respect to $w$. This can be shown for higher dimensional cases as well. 
\begin{theorem}[Differential of the closed-form]
Let $c(\Wno) = \sign (\Wno \x^*(\Wno))$ denote a sign vector. Then if  $c(\Wno)$ is a constant vector in an open neighbourhood containing $\Wno$, then the gradients of the closed-form expression  in \eqref{eq:closed_form} wrt $\Wno$ exist and the form of the differential \cite{minka_old_2000} is given by:\\
     \begin{align}
     & \partial \xopt =  - \proj_\nullspace{\Wzero}  \partial \Wpm\trans \s \label{eq:differential1}\\ 
      & \partial \xopt  = -(\C + \C\trans ) (\y - \beta \Wpm\trans \s) \label{eq:differential2} 
     \end{align}
where $\C=\Wzero^{\dagger} \partial\Wzero \proj_\nullspace{\Wzero}$, and  $\partial \xopt$, $\partial \Wzero$, and $\partial \Wpm$ are the differentials of $\xopt$,$\Wzero$, and $\Wpm$, respectively. 
\label{theorem2}
\end{theorem}

So based on Theorem~\ref{theorem2}, we present the final expression of the gradient of the upper level cost $\U(\Wno)$ with respect to $\Wno$.

\begin{lemma}
Let $\U(\Wno) = \frac{1}{2} \|\x^*(\Wno, \y_t) - \x_t \|_2^2 $ be the upper-level cost function that is smooth with respect to the intermediate reconstruction $\x^*(\Wno)$, then by the chain rule in matrix calculus, we derive an expression for the gradient of the cost $\U(\Wno)$ with respect to $\Wzero$ and $\Wpm$ (denoted by $\nabla_{\Wzero} \U$ and   $\nabla_{\Wpm} \U$) as\\
\begin{align}
    \nabla_{\Wpm} \U &= -\s \nabla_\xopt \U \trans    \proj_\nullspace{\Wzero}
    \label{eq:grad1}\\
    \nabla_{\Wzero} \U &= -  (\proj_\nullspace{\Wzero} 
    (\vec{q} \nabla_\xopt \U \trans  + \nabla_\xopt \U \vec{q}\trans)
    \Wzero^{\dagger})\trans, 
    \label{eq:except}
\end{align}
with $\vec{q} = \y_{t} -  \Wpm\trans \s$.
Here,  $\nabla_\xopt \U = (\xopt - \x_{t})$.
\end{lemma}

\section{BLORC Algorithm}
We propose an iterative Bilevel Learning Of $\ell_1$ Regularizers with Closed-form gradients (BLORC) algorithm for~\eqref{eq:bilevel}.
We perform minibatch gradient descent to learn the transform matrix $\Wno$ from supervised training pairs $(\x_{t},\y_{t})$. 
In all our experiments, we start from $\Wno_{0} = \vec{I}_{n \times n}$, the identity matrix. At the start of each epoch, the training pairs are randomly shuffled to remove any kind of dataset bias. Each sample in a batch is processed as follows.
\textbf{(a)} given the measurement $\y_{t}$ and current $\Wno$, the lower-level reconstruction problem is solved iteratively using ADMM to obtain an estimate of $\x_{t}^*(\Wno)$.  
\textbf{(b)} the sign vector $\sign (\Wno \x_{t}^*(\Wno))$ is obtained after hard-thresholding $\Wno \x_{t}^*(\Wno)$ with a small parameter $\gamma$. The sign vector is of paramount importance as it decides the row-split of $\Wno$ into $\Wzero$ and $\Wpm$. Hence, we run ADMM for a few thousands of iterations to ensure convergence of the sign pattern.
\textbf{(c)}  Then, using \eqref{eq:grad1} and \eqref{eq:except}, we 
obtain the
gradient of the upper level cost on a single training pair, i.e.,  $\nabla_{\Wno} \U$  which is the row-concatenation of  $\nabla_{\Wzero} \U$  and  $\nabla_{\Wpm} \U$. Averaging the gradients over the samples in the batch yields the minibatch gradient.  

At the end of each batch, we update the matrix $\Wno$ based on the learning rate $\alpha$ and the mini-batch gradient. The updated $\Wno$  is used in the next batch in step \textbf{(a)} above. 
Our method of gradient calculation can also be extended when the training pairs are image patches instead of 1D signals. For image denoising experiments, image patches of size $\sqrt{n} \times \sqrt{n}$ with an overlap stride of $r$ were extracted. Then the 2D patches were converted to 1D arrays as a pre-processing (first) step. In the end, the rows of the learned $\Wno$ were reshaped to look like convolutional filter patches. Except this first and last step, all the intermediate steps were the same for both 1D signals and image patches.

\section{Numerical Experiments}
To demonstrate how well BLORC learns the transform matrix $\Wno$, we perform denoising experiments on  synthetic 1D signals.
These synthetic 1D signals were chosen to be sparse with respect to a specific transform that also provides a baseline to compare our learned transforms with.

\vspace{-0.1in}
\subsection{Denoising experiments}
\begin{figure}
    \centering
    \includegraphics[width=0.4\textwidth]{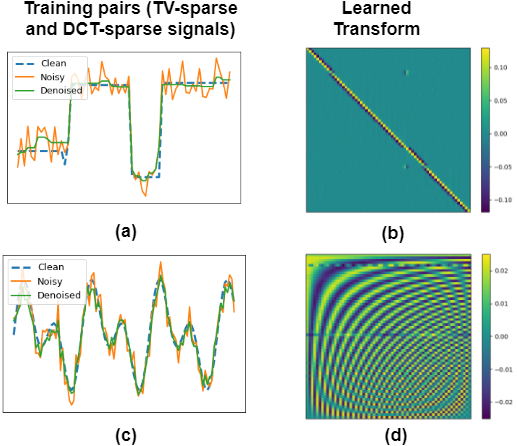}
    \caption{1D training pairs ($x_{t},y_{t}$) (left column) and the corresponding learned transform $\hat{\Wno}$ (right column).}
    \label{fig:lertran}
    \vspace{-0.1in}
\end{figure}

In our experiment, we generate $M= 4000$ training pairs $(\x_t,\y_t)$, where the $\x_t$'s are piece-wise constant signals of length $n=64$ (with peak value normalized to 1)  and $\y_t$'s are noisy versions with additive i.i.d. Gaussian noise with standard deviation $\sigma = 0.1$. Figure~\ref{fig:lertran}(a) shows a single pair of such $(\x_t,\y_t)$ . We perform minibatch gradient descent with batch-size $B=100$ and run the algorithm for $E=750$ epochs. The learning rate was chosen to be $\alpha = 10^{-4}$ and the sign threshold was $\gamma = 10^{-3}$. The learned transform is shown in Figure~\ref{fig:lertran}(b). We repeated the experiment with the same parameters but with 
$x_t$'s chosen as
smoothly varying signals of different harmonics
that are sparse in the discrete cosine transform (DCT) domain as in Figure~\ref{fig:lertran}(c).
The learned transform is row-rearranged such that it has maximum correlation with the 1D-DCT matrix and is shown in Figure~\ref{fig:lertran}~(d). What is interesting in the learned transforms of Figures~\ref{fig:lertran}(b) and \ref{fig:lertran}(d) is that they exactly capture the same intuition as the standard finite difference transform and the 1D-DCT transform, respectively, but in addition, they also have some novel features 
learned for the denoising task. Experimental results suggest that the learned transforms perform better than the standard transforms for denoising on a test-set. The average PSNR for 20 piecewise-constant test signals denoised using the learned $\Wno$ was 26.2 dB whereas that using the standard transform was 25.4 dB, with the PSNR of the noisy signals being 19.5 dB.   

\begin{figure}
    \centering
    \includegraphics[width=0.4\textwidth]{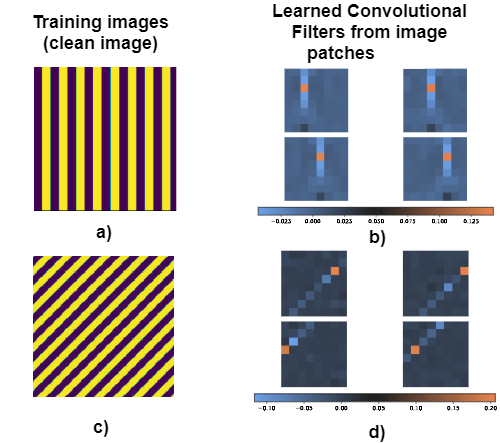}
    \caption{Training images (only clean) with a specific orientation (left) and learned convolutional filters (right).} 
    \label{fig:imagestrips}
    \vspace{-0.1in}
\end{figure}

Extending the BLORC algorithm to images or more specifically to image patches, we learn reasonable transforms as well. We chose images (normalized) of size $256 \times 256$ with directional patterns (vertical strips and diagonal strips) and generated their noisy versions with i.i.d. Gaussian noise with $\sigma =0.1$. We extracted image patches of size $8 \times 8$ with an overlap stride of $7$. The reason for choosing a large stride was to ensure that distinct training pairs are obtained. As a pre-processing step, the image patches of size $8 \times 8$ were vectorized to 
size $n=64$. All the parameters and the intermediate steps of the minibatch gradient descent are unchanged from the 1D signal experiment above. The learned convolutional filters in  Figs.~\ref{fig:imagestrips}(b) and \ref{fig:imagestrips}(d) look orthogonal to their image counterparts in Figs.~\ref{fig:imagestrips}(a) and  \ref{fig:imagestrips}(c), respectively, which is the expected output. 
In particular, these transforms are learned not to sparsify the training pairs but to minimize the gap between reconstructions and the ground truth.

\subsection{Accuracy of gradient computations}
\begin{figure}[hbt!]
    \centering
    \includegraphics[width=0.4\textwidth]{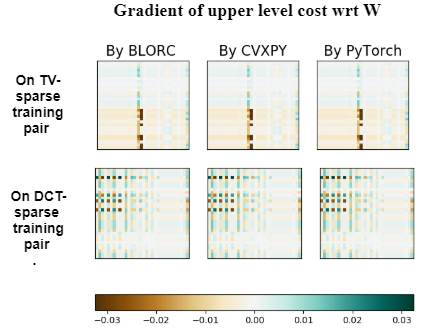}
    \caption{Gradient of upper level cost ($\nabla_\Wno \U$) for a single training pair $(x_{t},y_{t})$. }
    \label{fig:gradients}
    \vspace{-0.1in}
\end{figure}

While the BLORC algorithm uses an explicit form of the gradient,
in automatic differentiation approaches, the task is divided into a sequence of differentiable operations as computational graphs on which backpropagation is performed through the chain rule.  This division into a sequence of operations and calculating gradients for each node of the graph can take significant time, which can be bypassed if an explicit form of the gradient already exists that connects both the upper-level and the lower-level problem. This advantage in time can be crucial for larger datasets and batch-sizes. As a demonstration of this, we calculate and plot $\nabla_\Wno \U$ for a single training pair $(x_{t},y_{t})$ using three methods with the $\Wno = \vec{I}$ initialization in Figure~\ref{fig:gradients}. In the first method,  we use the direct expressions in~\eqref{eq:grad1} and~\eqref{eq:except} to get $\nabla_\Wno \U$, which we denote as "By BLORC" in Figure~\ref{fig:gradients}. For the second method, we used the CVXPY package to run an iteration of optimization over~\eqref{eq:bilevel} for the same training pair and obtained the gradient. Finally, as the third method, we used PyTorch to calculate the gradient of \emph{our closed-form expression}. As the baseline for our comparisons, we calculated the numerical gradient of the upper level problem in~\eqref{eq:bilevel} by noting the incremental change in the cost for incremental changes in each element of the matrix $\Wno$. The errors for the three methods in Table~\ref{table:compare} have been calculated with the 
ground truth value being set to the one from
the numerical gradient method. The analytical form of the gradient in
BLORC makes itt
faster and more accurate compared to automatic differentiation approaches as is evident from Table~\ref{table:compare}. In Figure~\ref{fig:gradients}, the gradient  $\nabla_\Wno \U$ calculated using all three methods encapsulates the feature of the single training pair $(x_{t},y_{t})$ but the one calculated using BLORC was more accurate overall as shown in Table~\ref{table:compare}.

\begin{table}[hbt!]
  \centering
  \renewcommand{\arraystretch}{1.2}
  \begin{tabular}{|p{1.2cm}|c|c|c|c|}
    \hline
    \multirow{2}{1cm}{\textbf{Gradient method}} & \multicolumn{2}{c|}{\textbf{$n=36$}} & \multicolumn{2}{c|}{\textbf{$n=64$}} \\
    \cline{2-5}
    & \textbf{Time (ms)} & \textbf{Error} & \textbf{Time (ms) } & \textbf{Error}\\
    \hline
    BLORC (ours) & 7.54 & 3.2e-09 &  12.65 & 5.13e-09 \\ \hline
    PyTorch  & 8.75 & 3.7e-09 & 14.03 & 5.35e-09 \ \\ \hline
    CVXPY & 17.85 & 4.8e-05  & 41.86 &  3.2e-05   \\ \hline
  \end{tabular}
   \caption{Time and error comparisions for Gradient calculation averaged over 100 different single training pairs $(x_{t},y_{t})$. }\label{table:compare}
   \vspace{-0.1in}
\end{table}

\section{Conclusion}
In this paper, we solve the bilevel problem of learning sparsity regularizers by analytically calculating the gradients of an implicit closed-form expression. The learned regularizers are optimal for the task at hand (here, denoising) and outperform known regularizers on test data for the task.
 Our mathematical analysis of bilevel reconstruction may lay the cornerstone for learning sparsifying (including deep) regularizers for 
inverse problems with 
general forward operators $\A$,
which is of major interest to the computational imaging community. 


\bibliographystyle{IEEEbib}
\bibliography{main}

\end{document}